\documentclass[conference]{IEEEtran}
\IEEEoverridecommandlockouts

\usepackage{cite}
\usepackage{amsmath,amssymb,amsfonts}
\usepackage{enumitem}
\usepackage{algorithmic}
\usepackage{graphicx}
\usepackage{textcomp}
\usepackage{xcolor}
\usepackage{tikz}
\usepackage[utf8]{inputenc}
\usepackage{newunicodechar}
\usepackage{booktabs}
\newunicodechar{①}{\textcircled{1}}

\def\BibTeX{{\rm B\kern-.05em{\sc i\kern-.025em b}\kern-.08em
    T\kern-.1667em\lower.7ex\hbox{E}\kern-.125emX}}
\begin{document}

\title{
Vision-Based Human Awareness Estimation for Enhanced Safety and Efficiency of AMRs in Industrial Warehouses
\thanks{© 2025 IEEE. Personal use of this material is permitted. Permission from IEEE must be obtained for all other uses, in any current or future media, including reprinting/republishing this material for advertising or promotional purposes, creating new collective works, for resale or redistribution to servers or lists, or reuse of any copyrighted component of this work in other works.}
}

\author{
\IEEEauthorblockN{
Maximilian Haug$^{a*}$,
Christian Stippel$^{b}$,
Lukas Pscherer$^{c}$,
Benjamin Schwendinger$^{a}$,\\
Ralph Hoch$^{c,d}$,
Angel Gaydarov$^{a}$,
Sebastian Schlund$^{a}$,
Thilo Sauter$^{d}$
}
\vspace{0.45em}\mbox{}
\IEEEauthorblockA{
$^{a}$\textit{Fraunhofer Austria Research GmbH}, 1040 Vienna, Austria\\
$^{b}$\textit{Computer Vision Lab}, TU Wien, 1040 Vienna, Austria\\
$^{c}$\textit{Digital Factory Vorarlberg GmbH}, 6850 Dornbirn, Austria\\
$^{d}$\textit{Institute of Computer Technology}, TU Wien, Vienna, Austria\\
$^{*}$corresponding author: maximilian.haug@fraunhofer.at}\\
\vspace{-2.5em}\mbox{}
}

\maketitle

\begin{abstract}
Ensuring human safety is of paramount importance in warehouse environments that feature mixed traffic of human workers and autonomous mobile robots (AMRs).
Current approaches often treat humans as generic dynamic obstacles, leading to conservative AMR behaviors like slowing down or detouring, even when workers are fully aware and capable of safely sharing space.
This paper presents a real-time vision-based method to estimate human awareness of an AMR using a single RGB camera.
We integrate state-of-the-art 3D human pose lifting with head orientation estimation to ascertain a human's position relative to the AMR and their viewing cone, thereby determining if the human is aware of the AMR.
The entire pipeline is validated using synthetically generated data within NVIDIA Isaac Sim, a robust physics-accurate robotics simulation environment.
Experimental results confirm that our system reliably detects human positions and their attention in real time, enabling AMRs to safely adapt their motion based on human awareness.
This enhancement is crucial for improving both safety and operational efficiency in industrial and factory automation settings.
\end{abstract}


\section{Introduction}
Employing autonomous mobile robots (AMRs) to transport goods in warehouses introduces challenges for human–robot interaction and safety.
Mixed human–robot traffic in warehouse aisles forces AMRs to trade off safety against efficiency: AMRs either slow down, stand still, or perform wide detours whenever a person enters their trajectory.  
Yet a detour is often unnecessary if the worker has noticed the robot, whereas a direct route can be dangerous when the worker remains unaware.
Traditional implementations treat humans as dynamic obstacles and respond by increasing the obstacle’s safety radius or replanning paths~\cite{kenk2019human}.
More recent work on human-aware navigation demonstrates that accounting for a person’s attention lets AMRs stay closer and still operate safely~\cite{cathcart2023proactive,hosseinzadeh2023toward,fischer2023collision, ansari2023exploring}.  
However, these studies focus mainly on optimizing the resulting trajectories and assume that each human’s awareness is already known.
In general, to the best of our knowledge, no existing system enables an AMR in a warehouse setting to visually detect if a human is paying attention to it.

\begin{figure}[t]
    \centering
    \includegraphics[width=\columnwidth]{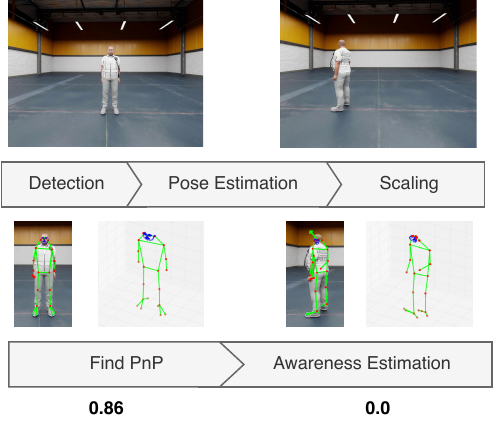}
    \caption{
        Overview of our awareness estimation pipeline. From a monocular RGB image stream, the system performs person detection, lifts keypoints to 3D, and estimates awareness using head orientation relative to the robot. The result is a 3D position and an awareness value for each person in the scene.
    }
    \label{fig:pipeline}
\end{figure}

Our work closes this gap with a vision pipeline that runs at 20 FPS on a consumer GPU and provides both the 3D position and an awareness indicator using only a RGB image stream as the input modality.
The core idea is to analyze visual cues of human attention, primarily body pose and head orientation, from the robot’s perspective.
Figure~\ref{fig:pipeline} shows that our pipeline receives an RGB image stream as input and performs following steps: (i)~YOLO person detection~\cite{varghese2024yolov8},  
(ii)~RTMW 3-D pose lifting~\cite{jiang2024rtmw} and  
(iii)~a geometry-based classifier that labels each persons awareness and 3D position relative to the AMR.
By simulating a warehouse scenario with a virtual human, we demonstrate that our framework can reliably estimate human awareness of the robot.
\\~\\
Our contribution can be summarized as follows:
\begin{itemize}
  \item A real-time 3D human detection algorithm that only requires RGB as a modality, inferring 3D.
  \item Introducing an end-to-end vision system for awareness detection in warehouse AMRs.
  \item An Isaac Sim warehouse simulation to evaluate our pipeline.
\end{itemize}

With our method, we enable ``situation aware'' navigation, where AMRs adjust their movements not only based on where people are, but also on their situational awareness.

\section{Related Work}
Recent work on industrial AMR deployments views humans primarily as dynamic obstacles.
Kenk et al.~\cite{kenk2019human} detect and track warehouse workers using onboard vision and integrate these detections into local costmaps for safe navigation.
Indri et al.~\cite{indri2020} extend this idea with online supervised global replanning.
Non-visual solutions, such as the LiDAR safety zones of Huang et al.~\cite{huang2021} enforce protective zones around operators, prompting the robot to halt or retreat whenever a person approaches.
While effective at preventing collisions, these approaches treat humans as passive obstacles and do not consider whether a person is aware of the robot.

Several studies indicate that accounting for human awareness can enhance navigation quality; however, most of these studies assume that awareness labels are provided externally.
Ansari et al.~\cite{ansari2023exploring} introduce a binary ``awareness'' flag into a crowd-navigation planner, reporting smoother, shorter trajectories when pedestrians are presumed to notice the robot.
However, they explicitly state that estimating awareness is ``out of scope''.
Paulin et al.~\cite{paulin2019} also assume the planner knows when a nearby person is looking at the robot, allowing for more assertive motion in tight spaces.
While they demonstrate the benefits of awareness-informed planning, they defer the challenge of perceiving awareness in real-time.

Finally, vision-based estimation of gaze and attention has mainly been explored in collaborative or social robotics scenarios rather than in warehouse AMRs~\cite{hanifi2024pipeline}.
Lavit Nicora et al.~\cite{lavit2024gaze}  implement real-time gaze detection, enabling a stationary cobot to initiate assembly steps only when the operator looks at it, enhancing timing and user comfort.
Di Stefano et al.~\cite{distefano2025} track eye gaze to guide a manipulator toward objects the human is attending to.
These studies confirm that head pose or eye-gaze cues can be extracted in real-time and mapped to adaptive robot behavior, but they are limited to fixed-base robots or desktop collaboration.
To date, no published system infers gaze-based awareness from an AMR's onboard camera in a warehouse and integrates this into the robot’s navigation stack.
Addressing this gap motivates the proposed vision-based awareness pipeline.

\section{Simulation Environment}
To implement and evaluate the proposed vision-based AMR-human awareness system, NVIDIA Omniverse and its robotics extension are used to build a simulation environment.

\subsection{NVIDIA Omniverse and Isaac Sim}
NVIDIA Omniverse is a development platform for real-time, photorealistic 3D simulation, built on the Universal Scene Description framework and RTX ray tracing. NVIDIA Isaac Sim~\cite{nvidia_isaac_sim_manual}, built on Omniverse, provides a physics-accurate robotics simulation environment with sensor modeling, synthetic data generation, and support for human characters and environmental assets. We use Isaac Sim as our core simulation engine to prototype and validate an AMR in a realistic virtual environment.

\subsection{Simulation Setup}
A realistic warehouse scene is built in Isaac Sim (using the NVIDIA warehouse environment extension) to serve as a test environment.
The Nova Carter robot, an autonomous mobile robot platform, is imported as our AMR model.
A virtual human character (a warehouse worker) is added into the scene and animated to walk across the warehouse floor while periodically rotating its head (mimicking natural gaze shifts).
To approximate the perspective of a future forklift-mounted camera, the camera is mounted on a virtual post elevated above the Nova Carter robot’s base.
The camera that is used on the AMR is an RGB pinhole camera to avoid lens distortions.
A second virtual RGB pinhole camera is attached to the head of the human character to capture the human’s field of view (FOV).
During the simulation, the AMR drives forward toward the path of the moving human, while the human walks across the scene, creating a dynamic scene.
This interaction is captured and serves as input to our vision processing pipeline.

\section{Awareness Estimation}
\label{sec:method}

The awareness-estimation pipeline runs fully on the AMR's RGB camera stream.
It comprises four stages: person detection, 2D/3D keypoint extraction, head pose recovery, and awareness scoring.
First, each frame is processed by the system using MMDetection with a YOLO backbone to detect the person in the scene~\cite{varghese2024yolov8}.
The highest-confidence human bounding box $\mathcal{B}$ is selected and the image is cropped to $\mathcal{B}$ to reduce pose-estimation latency.
The cropped region is fed to RTM3D in MMPose~\cite{mmpose2020}, which outputs 
\begin{itemize}
    \item 2D keypoints $\bigl\{\mathbf u_i\bigr\}_{i=1}^{K}$ in pixel space and
    \item canonical camera 3D keypoints $\bigl\{\tilde{\mathbf p}_i\bigr\}_{i=1}^{K}$
\end{itemize}
where $K$ denotes the total number of anatomical keypoints for one human as determined by the pose estimation model.
Because the canonical scale is unitless, each limb length is rescaled to metric units using anthropometric averages, yielding metric 3D keypoints $\mathbf{p}_i =\mathcal{s}_i\tilde{\mathbf{p}_i}$ where $\mathcal{s}_i$ is the limb-specific scale factor (e.g, shoulder span $\approx 0.45$m).

Furthermore, the coordinate origin is transformed to the nose tip ($\mathbf{p}_{nose}=0$) and the positive $\mathcal{y}$-axis is aligned with the forward direction (gaze direction) of the head.

An initial gaze vector 
\[
\hat{\mathbf g}_0 \;=\;
\frac{\mathbf p_{\text{left-ear}}-\mathbf p_{\text{right-ear}}}
     {\bigl\|\mathbf p_{\text{left-ear}}-\mathbf p_{\text{right-ear}}\bigr\|}
\;\times\;
\frac{\mathbf p_{\text{nose}}-\mathbf p_{\text{neck}}}
     {\bigl\|\mathbf p_{\text{nose}}-\mathbf p_{\text{neck}}\bigr\|}.
\]

is orthogonalized to yield a right-handed head frame with $\hat{\mathbf{x}}$, $\hat{\mathbf{y}}$, and $\hat{\mathbf{z}}$

\begin{align*}
\hat{\mathbf x} \;&=\; \hat{\mathbf y}\,\times\,\hat{\mathbf z},\\
\hat{\mathbf y} \;&=\;
\frac{\hat{\mathbf g}_0}{\|\hat{\mathbf g}_0\|},\\
\hat{\mathbf z} \;&=\;
\frac{\mathbf p_{\text{right-ear}}-\mathbf p_{\text{left-ear}}}
     {\bigl\|\mathbf p_{\text{right-ear}}-\mathbf p_{\text{left-ear}}\bigr\|}.
\end{align*}

with $\hat{\mathbf y}$ pointing forward.
The transformation between the head pose and the  camera frame is obtained by solving the Perspective-n-Point correspondence (PnP) iteratively over a subset of head keypoints by minimizing the reprojection error:

\[
\bigl(\mathbf r_{\text{head}},\,\mathbf t_{\text{head}}\bigr)
\;=\;
\operatorname*{arg\,min}_{\mathbf r,\mathbf t}
\;\sum_{i\in\mathcal H}
\Bigl\|
\pi\!\bigl(\,\mathbf R(\mathbf r)\,\mathbf p_i + \mathbf t\bigr)
-\mathbf u_i
\Bigr\|^{2},
\]

where $\mathbf{u}_i$ are the 2D image keypoints, $\mathbf{p}_i$ are the corresponding 3D metric keypoints, $\mathbf{R(r)}$ is the matrix description of the rotation vector $\mathbf{r}$, $\pi(\cdot)$ is the camera projection function, and $\mathcal{H}$ denotes the set of head keypoints used in PnP optimization.
The optimization is performed with OpenCV, returning the rotation vector $\mathbf{r}$ and translation $\mathbf{t}$ that locate the head pose and gaze direction of the human with respect to the camera coordinate system.

From the estimated head pose, we construct a virtual ``attention cone'' to model the person’s FOV with configurable angle $\theta_\text{FOV}$ (e.g., representing human peripheral vision limits).
Let the cone apex $\mathbf{c}\in\mathbb{R}^3$ be the center of the approximation of the detected face, let the unit gaze vector be $\hat{\mathbf{g}}$, describing the aligned central axis of the cone with the forward direction of the head, and let $\mathbf{p}_a\in\mathbb{R}^3$ denote to the position of the AMR in the same frame.
The signed axial distance along the gaze axis is given by
 
\begin{equation}\label{eq:1}
    d = (\mathbf{p}_a -\mathbf{c})^\mathsf{T} \hat{\mathbf{g}}
\end{equation} 

which describes the scalar projection of the AMR vector onto the unit gaze vector. 
If the AMR’s position lies within the volume of this cone (i.e., the angle between the person’s gaze direction and the vector toward the AMR is below $\theta_\text{FOV}/2$), then the AMR is considered visible to the person.
To quantify the level of awareness on a continuous scale, we calculate the normalized minimal offset of the AMR from the center of the cone.
In practice, this is done by finding the shortest distance from the AMR to the central axis of the cone, which is given by 

\begin{equation}\label{eq:2}
 d_{\perp} = \bigl\|\,(\mathbf{p}_{a}-\mathbf{c}) - d\,\hat{\mathbf{g}}\,\bigr\|
\end{equation} 

and dividing it by the radius of the cone at that distance from the head. Whereby the cone radius at depth $d$ is \mbox{$r(d) = d\tan \mathbf{\theta}_{FOV}$}.
This yields the continuous awareness score \mbox{$\alpha\in[0, 1]$}:
\begin{equation}\label{eq:3}
\alpha =
\begin{cases}
1-\dfrac{d_\perp}{r(d)}, &\text{if }d>0\text{ and }d_\perp\le r(d),\\
0, &\text{otherwise}.
\end{cases}
\end{equation}

The term $\dfrac{d_\perp}{r(d)}$ represents the fractional radial displacement of the AMR from the central axis of the cone.
Thus, $\alpha = 1$ indicates the AMR is directly in the person’s line of sight (center of the gaze cone) and $\alpha = 0$ indicates the AMR is outside the FOV.
Intermediate values $0 < \alpha < 1$ correspond to the AMR falling within the FOV but outside the center, with higher values for $\alpha$ for more central positions.

Equations \eqref{eq:1}–\eqref{eq:3} are evaluated for every video frame, producing a real-time awareness signal $\alpha(t)$ that reflects moment-to-moment variations in head orientation and relative robot–human geometry.

\section{Experimental Results}
\label{sec:experiments}

\begin{figure*}[t]
  \centering
  \includegraphics[width=\textwidth]{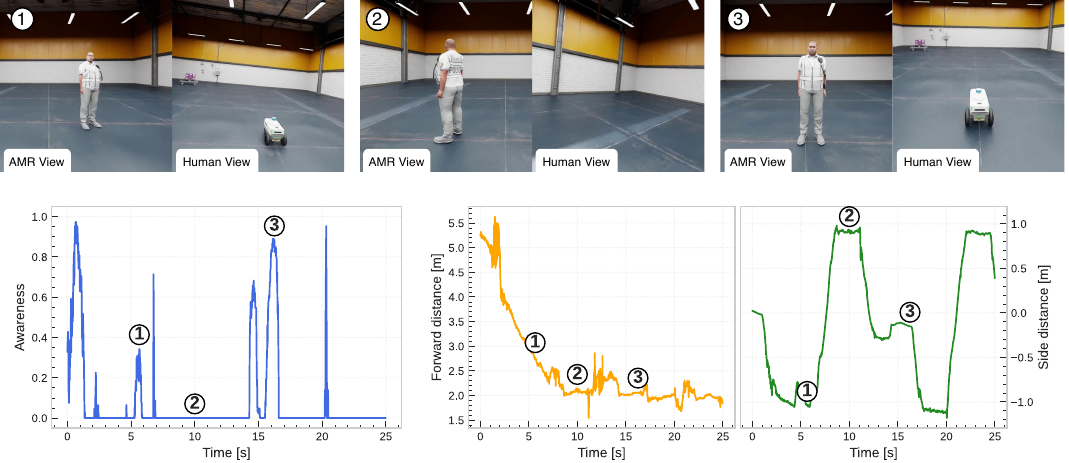}
  \caption{\textbf{Awareness–distance profiles in a 25\,s encounter.}
           \emph{Top:} Three representative image pairs show the scene from the
           AMR’s camera (left in each pair) and the human’s perspective
           (right), captured at the time stamps marked (1)–(3).
           \emph{Bottom-left:} Our pipeline’s continuous awareness score
           (blue). Peaks coincide with moments when the human glances toward the robot.  
           \emph{Bottom-middle:} Forward distance (orange) between the human and the AMR along the robot’s heading.  
           \emph{Bottom-right:} Lateral (side) distance (green), where positive values indicate the human is to the robot’s right.
           Together, the plots illustrate how awareness spikes precede, and therefore could trigger, adaptive speed or warning behaviours.
           }
  \label{fig:aware_distance}
  \vspace{-2ex}
\end{figure*}


Figure~\ref{fig:aware_distance} shows one representative 25-second encounter at 60 fps with 1500 frames between the Nova Carter AMR and a virtual warehouse worker.
While the robot drives straight ahead, the human walks laterally across its path.

Three signals are plotted over time:
\begin{itemize}
  \item \textbf{Awareness score} $\alpha(t)$ (blue, left):
        the pipeline output, normalized to $[0,1]$.
  \item \textbf{Forward distance} $d_{\mathrm{fwd}}(t)$ (orange, center):
        Euclidean distance along the robot’s heading.
  \item \textbf{Side distance} $d_{\mathrm{lat}}(t)$ (green, right):
        lateral offset; \mbox{$d_{\mathrm{lat}}>0$} means the human is on the
        robot’s left.
\end{itemize}

Three paired images (1)–(3) provide qualitative ground truth at characteristic times: the left shows the AMR’s camera view; the right shows the human-mounted eye camera. \\
\textit{Cases:}
    \begin{enumerate}[label={(\arabic*)}]
        \item Robot is located on the edge of the FOV of the human: \textbf{Partial awareness}
        \item Robot is outside the FOV: \textbf{No awareness}
        \item Robot centered in gaze: \textbf{Full awareness}
    \end{enumerate}

\begin{table}[h]
    \centering
    \caption{Qualitative cases highlighted in Fig.~\ref{fig:aware_distance}.}
    \label{tab:case_summary}
    \resizebox{\linewidth}{!}{
    \begin{tabular}{c c c c c}
        \hline
        \textbf{Case} 
        & \textbf{Time (s)} 
        & \textbf{$\alpha(t)$} 
        & \textbf{$d_{\mathrm{fwd}}(t)\,\mathrm{[m]}$} 
        & \textbf{$d_{\mathrm{lat}}(t)\,\mathrm{[m]}$} \\
        \hline
        (1) & 5.65  & 0.34 & 2.77 & $-1.04$ \\
        (2) & 10.00 & 0.00 & 2.13 & \phantom{$-$}0.92 \\
        (3) & 16.25 & 0.88 & 2.05 & $-0.15$ \\
        \hline
    \end{tabular}}
\end{table}

Peaks in $\alpha(t)$ at cases (1) and (3) align with moments when the eye-camera confirms that the human gazes at the robot. Zero awareness values appear when the human looks away from the robot, as seen in case (2). The human is completely unaware of the AMR.
These preliminary results demonstrate that our vision-based pipeline delivers timely, geometry-aware awareness signals suitable for future speed- or warning-adaptive navigation.

\section{Conclusion}
This paper introduces a vision-based pipeline for inferring human awareness of an approaching AMR using only a single onboard camera.
By combining real-time person detection, 3D keypoint extraction, and head-pose analysis, the system estimates both the person’s 3D position and a continuous ``awareness score'', validated in NVIDIA Isaac Sim.

The ``awareness score'' enables AMRs to adapt navigation dynamically, slowing only if people nearby are unaware, while maintaining efficiency when they are attentive.
Although our synthetic experiments show feasibility, limitations remain.
We currently handle single-person cases, approximate gaze from head orientation alone, and rely on simulated RGB imagery.
Future work will address real-world conditions, integrate eye-gaze estimation, and validate awareness detection with human participants.
Furthermore, our awareness score can be integrated directly into any navigation stack that deals with dynamic obstacles by adjusting the 3D bounding box based on the awareness score.

\section{Acknowledgement}
This research was conducted within the project AUTARK, funded by the Federal Ministry of Innovation, Mobility and Infrastructure (BMIMI) via the Austrian Research Promotion Agency (FFG) under contract no. FO999922723, within the programme "\textit{Schlüsseltechnologien im produktionsnahen Umfeld}".
\bibliographystyle{IEEEtran}

\bibliography{etfa_autark}
\end{document}